\documentclass[twoside]{article}

\usepackage{PRIMEarxiv}

\usepackage[utf8]{inputenc} 
\usepackage[T1]{fontenc}    
\usepackage{hyperref}       
\usepackage{url}            
\usepackage{booktabs}       
\usepackage{amsfonts}       
\usepackage{nicefrac}       
\usepackage{microtype}      
\usepackage{lipsum}
\usepackage{adjustbox}
\usepackage{fancyhdr}       
\usepackage{graphicx}       
\graphicspath{{media/}}     
\usepackage{multirow}

\usepackage{newunicodechar}

\usepackage[most]{tcolorbox}
\usepackage{enumitem}
\usepackage[table]{xcolor}  
\usepackage[most]{tcolorbox}
\usepackage{xcolor}
\usepackage[normalem]{ulem}
\usepackage{pifont}

\newtcolorbox{promptbox}[1]{
    enhanced,
    breakable,
    width=\columnwidth,
    title=\textbf{#1},
    colback=gray!3,
    colframe=gray!40,
    colbacktitle=gray!10,
    coltitle=black,
    fonttitle=\small,
    fontupper=\footnotesize\ttfamily,
    boxrule=0.5pt,
    arc=2pt,
    left=5pt,
    right=5pt,
    top=5pt,
    bottom=5pt
}

\title{Learning Transformation-Induced Representational Differences for Multilingual AI-Generated Text Detection}

\author{
Sujit Kumar\\
  A*STAR Institute of Advanced\\ Intelligence and Computing\\ (A*STAR IAIC) 
  Singapore\\ 
\And  
   Shifali Agrahari \\
  Department of CSE, \\
   Indian Institute of Technology Guwahati \\
  Assam, India\\
  \texttt{a.shifali@iitg.ac.in} \\   
  \And
  Sanasam Ranbir Singh \\
   Department of CSE, \\
   Indian Institute of Technology Guwahati \\
  Assam, India\\
}

\begin{document}
\maketitle

\begin{abstract}
As Large Language Models (LLMs) continue to advance and become increasingly multilingual, distinguishing between human-authored and LLM-generated text is crucial for preserving trust and integrity in information dissemination, public discourse, and digital ecosystems. However, existing studies on AI-generated text detection have predominantly focused on English, leaving significant research gaps in low-resource languages. This study proposes two novel methods, Multilingual Text Recognition Using Semantic Transformations (M-TRUST) and Residual Drift for AI-Text Detection (RED), for detecting multilingual AI-generated text. Our proposed methods, M-TRUST  and RED, leverage semantically preserving text transformations as stress tests to distinguish AI-generated from human-authored text. Meanwhile, the M-TRUST  model employs multilingual XLM-RoBERTa to explicitly attend to the original and transformed variants, enabling it to capture token-level divergences. In contrast, the RED method measures residual differences between the embeddings of the original and transformed texts to quantify the difference between the original and transformed texts. We conduct our experiments and ablation studies over 13 datasets spanning across 11  linguistically diverse languages, including low-resource
languages. Experimental results and ablation studies demonstrate that the proposed methods outperform existing state-of-the-art approaches across datasets spanning linguistically diverse languages. Furthermore, the proposed approaches effectively distinguished AI-generated text and human-authored text across linguistically diverse languages, including low-resource languages, and demonstrated generalisation to unseen text generators and text transformation models.

\end{abstract}

\section{Introduction}
The rapid development of Large Language Models (LLMs) has driven substantial progress across a range of natural language generation tasks, such as writing news articles~\cite{yanagi2020fake,singh2024clustercore,kumar2026prompt}, generating stories~\cite{yuan2022wordcraft}, generating code~\cite{agrahari2026osint} and generating reviews~\cite {salminen2022creating, agrahari2025can}. As LLMs become increasingly multilingual \cite{lai2023chatgpt, agrahari2025osint, agrahari2025essaydetect} and widely accessible, the associated risks have also increased. 
The misuse of LLM-generated content can contribute to the proliferation of fake news and misinformation~\cite{wang2022fake,kumar2025fake,kumar2024headline}, scams by generating convincing malicious emails, and facilitation of academic dishonesty \cite{mirsky2021creation, zellers2019defending, perkins2023academic}. These threats underscore the urgent need for robust detection methods to distinguish human-written from LLM-generated text across languages and domains \cite{guo2023close, su2023detectllm}. 

\begin{figure*}[ht]
    \centering
    \includegraphics[width=\linewidth]{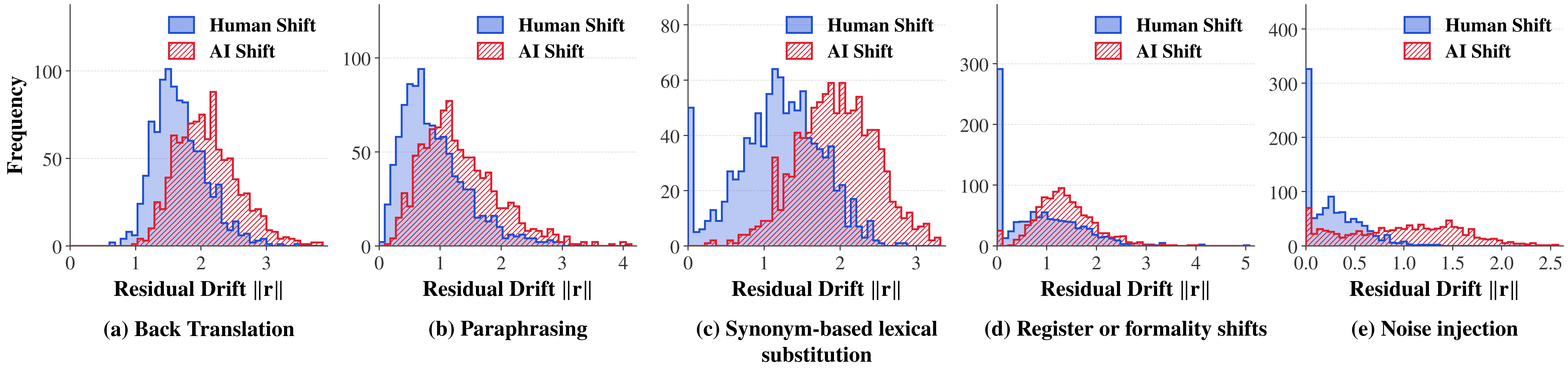}
    \caption{
    Visualization of \textbf{Residual Drift Distributions} for human-written and LLM-generated texts across five stylistic transformations: \textit{translation}, \textit{paraphrase}, \textit{formality}, \textit{lexical}, and \textit{noise injection}. 
    Each subplot shows the histogram of residual drift magnitudes ($\|r\|$), computed as described in Equations~(5)--(7). 
    The \textcolor{blue}{blue bars} represent human-authored texts, and the \textcolor{red}{red bars} denote LLM-generated texts. 
    Overall, human-written texts exhibit broader and right-shifted distributions, indicating higher stylistic variability, whereas LLM-generated texts show narrower distributions, reflecting greater stylistic stability.}
    \label{fig:first_image}
    
\end{figure*} 
In the literature, LLM-generated text detection approaches can be broadly categorised into three groups: (i) supervised methods, (ii) zero-shot methods, and (iii) transformation-based methods. Supervised methods~\cite{solaiman2019release, fagni2021tweepfake, su2023detectllm, ahmed2018detecting, liyanage2024detecting, liu2023argugpt} for LLM-generated text detection are trained on labelled datasets containing both LLM-generated and human-authored text. However, supervised methods for LLM-generated text detection perform well in-domain, but their performance degrades significantly under out-of-distribution (OOD) settings~\cite{wang2024m4, agrahari2025tracing, agrahari2025random}. Zero-shot methods leverage LLM-generated logits~\cite{mao2024raidar, ma2024zero, yu2024dpic, park2025dart, nguyen2024simllm, gehrmann2019gltr, mitchell2023detectgpt} to extract various perturbation and probability-based metrics, and then build threshold-based classifiers.
Transformation-based methods RAIDAR~\cite{mao2024raidar} and
LLaMA-L2R~\cite{hao-etal-2025-learning}, exploit differences
in LLM-based rewriting behaviour between human-authored
and AI-generated texts, whereas
SimLLM~\cite{nguyen2024simllm} leverages the similarity
between input texts and their LLM-regenerated counterparts. Despite their effectiveness, existing transformation-based detection approaches suffer from several limitations: (i) their reliance on surface-level edit distances or aggregate similarity measures may inadequately capture fine-grained contextual divergences and subtle semantic changes induced by text transformations; (ii) they are potentially vulnerable to rewriting-based adversarial attacks, which may weaken the discriminative signals used to distinguish human-authored from LLM-generated texts; (iii) establishing reliable edit-distance-based decision boundaries is challenging, as language-specific lexical and morphological characteristics may influence textual modification patterns across languages and domains; and (iv) their predominant reliance on rewriting or regeneration limits the exploration of discriminative signals arising from diverse meaning-preserving linguistic transformations. Furthermore, their effectiveness across linguistically diverse and low-resource languages remains largely underexplored. Furthermore, as highlighted by study~\cite{macko2023multitude, agrahari2026bhaid}, multilingual AI-generated text detection continues to face major challenges, including (i) the absence of language-agnostic detection strategies that generalise across diverse languages and scripts~\cite{macko2023multitude}, (ii) persistent performance gaps between high-resource and low-resource languages~\cite{macko2025multisocial}, and (iii) the lack of standardised multilingual benchmarks enabling fair and reproducible evaluation~\cite{macko2023multitude}.

To overcome the aforementioned limitations, we propose two novel methods for LLM-generated text detection across diverse languages : Multilingual Text Recognition Using Semantic Transformations \textbf{(M-TRUST)} and Residual Drift for AI-generated Text Detection \textbf{(RED)}. Our proposed methods, M-TRUST and RED, first apply transformations to text using LLMs, then learn semantic and lexical shifts between the original and transformed texts to distinguish human-written from LLM-generated content. Our proposed methods adopt monolingual paraphrasing, synonym-based lexical substitutions, register or formality shifts, and noise-injection transformations using LLMs. These transformations expose subtle inconsistencies and stylistic rigidities in LLM-generated text, enabling the models to capture robust semantic and lexical cues for AI-generated text detection. Our key contributions can be summarised as follows: (i) We propose two novel frameworks,  Multilingual AI-Generated Text Recognition Using
Semantics-Preserving Transformations~\textbf{(M-TRUST)} and Residual Drift for AI-generated Text Detection (RED), for AI-generated text detection. 
(ii) We investigate multiple text transformation strategies, including monolingual paraphrasing, synonym-based lexical substitution, register and formality shifts, and noise injection, for detecting AI-generated content across eleven languages. (iii) Through several ablation studies over thirteen datasets from eleven languages, we demonstrate that our proposed methods, M-TRUST and RED, are highly effective at detecting LLM-generated text across multiple languages, including low-resource languages.  

\section{Related Work}
In literature studies~\cite{wu2025survey, xu2025advancements, agrahari2024text} provide comprehensive reviews of datasets and methodologies for AI-generated content detection. Existing approaches to AI-generated content detection include supervised learning, zero-shot detection, and transformation-based techniques. Supervised learning-based methods train models on labelled human-authored and AI-generated text to distinguish between the two~\cite{xiao2025human, wang2025genai, solaiman2019release, fagni2021tweepfake, su2023detectllm, ahmed2018detecting, liyanage2024detecting, liu2023argugpt}.  Supervised methods~\cite{agrahari2026more, agrahari2025osint} perform well in-domain but show significant performance degradation under out-of-distribution (OOD) scenarios \cite{wang2024m4}, particularly when evaluated on unseen models, languages, or domains. Zero-shot methods, such as GLTR~\cite {gehrmann2019gltr}, extract statistical features such as log-probability and entropy; DetectGPT \cite{mitchell2023detectgpt} employs token perturbation to compute log-probability.  Despite their effectiveness, zero-shot detection methods often rely on statistical characteristics, model-dependent probability estimates, or similarity-based measures, which may not generalise reliably across different text generators, domains, and languages. SimLLM~\cite{nguyen2024simllm} leverages semantic similarity between the original text and its LLM-generated counterpart, while DART~\cite{park2025dart} examines semantic differences between the original and rephrased text.

Transformation-based approach RAIDAR~\cite{mao2024raidar} exploits differences in rewriting behaviour between human-authored and AI-generated text, whereas LLaMA-L2R~\cite{hao-etal-2025-learning} adopts a learning-based approach to rewriting-oriented detection. TRIDENT~\cite{agrahari2026trident} introduces a round-trip translation-based detection framework that exploits semantic consistency between original and translated-back texts. However, existing transformation-based methods face several limitations: (i) \textit{Limited contextual sensitivity}: reliance on edit-distance or similarity-based measures may overlook fine-grained contextual divergences and subtle semantic variations induced by text transformations. (ii) \textit{Vulnerability to adversarial rewriting}: LLM-based paraphrasing may reduce the discriminative differences between human-authored and AI-generated text, potentially undermining detection reliability. (iii) \textit{Threshold sensitivity}: edit-distance-based detection requires reliable decision thresholds, which may vary across languages, domains, and text characteristics. (iv) \textit{Limited transformation diversity}: predominant reliance on rewriting restricts the exploration of discriminative signals arising from diverse meaning-preserving linguistic transformations. (v) \textit{Limited generalizability}: the transferability of transformation-based detection across different text generators, transformation strategies, and linguistically diverse languages remains insufficiently explored, particularly given language-specific lexical and morphological variations.

Motivated by these limitations, we propose two complementary AI-generated content detection models: M-TRUST, which captures fine-grained token-level contextual divergences between original and transformed texts, and RED, which quantifies representation-level semantic drift to identify subtle changes induced by linguistic transformations. Both M-TRUST and RED leverage diverse meaning-preserving transformations to distinguish human-authored from AI-generated text, while enabling systematic evaluation across different text generators, transformation strategies, and languages. We further examine the robustness and generalizability of these signals across different text generators, transformation strategies, and linguistically diverse languages.

\begin{figure*}[]
    \centering
\includegraphics[width=\linewidth]{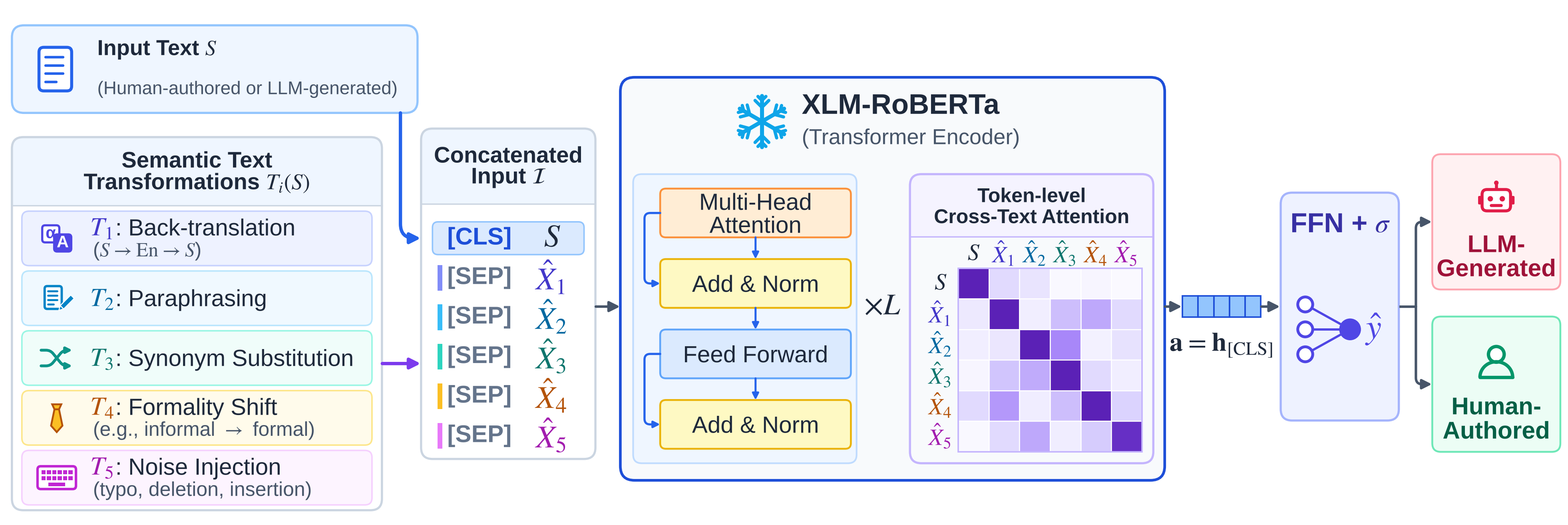} 
\caption{Overview of the proposed M-TRUST framework, where five semantics-preserving transformations are generated using LLMs and concatenated with the original input text. The combined text is processed through XLM-RoBERTa to capture token-level cross-text interactions and semantic variations for distinguishing human-authored from LLM-generated text.}
    \label{fig:mtrust}
\end{figure*}
\section{Proposed method}
This study proposes two novel methods, Multilingual Text Recognition Using Semantic Transformations \textit{M-TRUST} and Residual Drift for AI-text Detection \textit{RED}, for detecting LLM-generated text by exploiting the invariance and variability of LLM-generated and human-authored text under diverse transformations. The prime motivation behind our proposed methods is that human-authored and LLM-generated texts respond differently to meaning-preserving transformations. Specifically, LLM-generated text exhibits greater consistency with its transformed variants, whereas human-authored text shows larger variations. These differences  provide an important cue for distinguishing human-authored from LLM-generated text. As shown in Figure~\ref{fig:first_image}, human-authored text undergoes greater changes after transformations, indicating higher residual drift compared to the more stable behaviour of LLM-generated text. The distributions of drift values for human text are wider and shifted towards higher magnitudes, particularly under translation and rewriting transformations. In contrast, LLM-generated text demonstrates relatively smaller shifts, with narrower distributions clustered around lower drift values.  This indicates that LLMs tend to maintain their characteristic writing style even after transformation, leading to greater stability. In contrast, human-authored text exhibits higher variability after transformation, driven by richer linguistic diversity, contextual nuance, and individual writing styles. Consequently, we conclude that human-authored text undergoes more noticeable semantic drift under transformations, whereas LLM-generated text remains unnaturally consistent, following its own stylistic pattern.\par
We consider five different text transformations over input texts: (i) \textit{One-round back-translation}: Through prompt instructions, we perform one-round back-translation via English as a pivot language, where the source text is translated into English and then back to its source language. Our prompt instruction for one-round back-translation is as follows: \textit{Prompt:} \textit{Translate the given text into English and then translate it back into Hindi. Ensure meaning is preserved while allowing natural variation.} (ii) \textit{Paraphrasing}: We apply the paraphrasing transformation by paraphrasing the text into the same language through LLMs with the help of prompt instructions. \textit{Prompt for Paraphrasing transformations :} \textit{Rewrite the following sentence differently while preserving its meaning.} (iii) \textit{Synonym-based lexical substitution} :  Synonym-based lexical substitution replaces selected words in a sentence with semantically equivalent synonyms to generate meaning-preserving lexical variants. Our prompt instruction to LLMs, Synonym-based lexical substitution, is as follows: \textit{Prompt: Replace some words in the given text with appropriate synonyms while keeping the overall meaning intact.} (iv) \textit{ Register or formality shifts}: Modify the stylistic tone of text by converting the text into formal through prompt instructions with LLMs while preserving semantic content. \textit{Prompt: Rewrite the following text in a more formal style while preserving its meaning.} (v) \textit{Noise injection} :  Introduces minor random perturbations into text, such as token deletion, insertion, or replacement, to simulate natural text variability while preserving meaning. \textit{Prompt: Introduce minor variations such as spelling mistakes, typographical errors, or small grammatical shifts, while keeping the text understandable.}\par

Given a source text $\mathcal{S}$, which may be authored either by a human or an LLM, we first obtain a set of transformed texts 
$\acute{\mathcal{X}}_1, \; \acute{\mathcal{X}}_2, \; \acute{\mathcal{X}}_3, \; \acute{\mathcal{X}}_4, \; $ $\acute{\mathcal{X}}_5$ 
by applying  transformation functions 
$T_1(\mathcal{S}), \; T_2(\mathcal{S}), \; T_3(\mathcal{S}), \; T_4(\mathcal{S}),$ \; $ T_5(\mathcal{S})$ 
over the input text $\mathcal{S}$ through Equations as defined below.
\begin{equation}
\label{Eq:transformation}
     \acute{\mathcal{X}}_1, \; \acute{\mathcal{X}}_2,\; \acute{\mathcal{X}}_3,\; \acute{\mathcal{X}}_4,\; \acute{\mathcal{X}}_5\; \text{=} \; T_1(\mathcal{S}), \; T_2(\mathcal{S}),\; T_3(\mathcal{S}),\; T_4(\mathcal{S}),\; T_5(\mathcal{S})
\end{equation}
Where $T_1(\mathcal{S})$, $T_2(\mathcal{S})$, $T_3(\mathcal{S})$, $T_4(\mathcal{S})$, and $T_5(\mathcal{S})$ denote one-round back-translation, paraphrasing, synonym-based lexical substitution, register or formality shifts, and noise injection transformations, respectively.

\subsection{Multilingual Text Recognition Using Semantic Transformations  (M-TRUST )}
Figure~\ref{fig:mtrust} presents the working of our proposed framework M-TRUST. Given  transformed texts  $\acute{\mathcal{X}}_1, \; \acute{\mathcal{X}}_2, \; \acute{\mathcal{X}}_3, \; \acute{\mathcal{X}}_4, \; \acute{\mathcal{X}}_5$ of a source text $\mathcal{S}$ we  concatenate all transformed text with $\mathcal{S}$ as defined below. 
\begin{equation}
\label{lab:inp_text}
    \text{I} \;\text{=}\; [\texttt{[CLS]} \; \mathcal{S} \; \texttt{[SEP]} \; \acute{\mathcal{X}}_1, \; \texttt{[SEP]} \; \acute{\mathcal{X}}_2 \;\texttt{[SEP]} \; \acute{\mathcal{X}}_3 \;\texttt{[SEP]} \; \acute{\mathcal{X}}_4 \; \texttt{[SEP]}\; \acute{\mathcal{X}}_5].
\end{equation}
Next, we feed concatenated input text \text{I} obtained through  Equations~\ref{lab:inp_text} to a 	XLM-RoBERTa to obtain an encoded representation $\mathbf{a}$ of \text{I}. The primary motivation behind encoding the source text and its transformed variants using a XLM-RoBERTa is explicitly capturing token-level divergences between the original and transformed text. The multi-head attention component of the XLM-RoBERTa facilitates this process by jointly attending to corresponding tokens across the original text and the transformed text, enabling the encoder model to highlight fine-grained lexical and contextual mismatches that distinguish human-authored text from LLM-generated text. Intuitively, if the transformed text remains close in meaning and structure to the original, the corresponding tokens exhibit higher attention weights, indicating more substantial semantic alignment. In contrast, when the transformed text diverges substantially from the original, the attention weights become weaker and more scattered, reflecting semantic drift. Such variations in attention distributions provide a fine-grained signal for detecting stylistic rigidity in LLM-generated text and linguistic flexibility in human-authored text. 

Subsequently, the encoded the representation $\mathbf{a}$ of \text{I} is pass through  a feedforward neural network followed by a sigmoid function for classification of $\mathcal{S}$ into LLMs generated or human-authored by following the equation as defined below. Subsequently, the loss function is defined as shown in Equation~\ref{eq:clas_loss} below.

\begin{equation}
\label{eq:clas_loss}
\hat{y} = \sigma(\mathbf{W}_a \cdot \mathbf{a}  + \mathbf{b}_a)
\end{equation}
where $\mathbf{W}_a$ and $\mathbf{b}_a$ are learnable  parameters, $\mathbf{a}$ is encoded representation of  \text{I},  and $\sigma(\cdot)$ denotes the sigmoid activation. The model is optimised using the binary cross-entropy (BCE) loss between the predicted probabilities and the ground-truth labels. 

\begin{figure*}[]
    \centering
    \includegraphics[width=\linewidth]{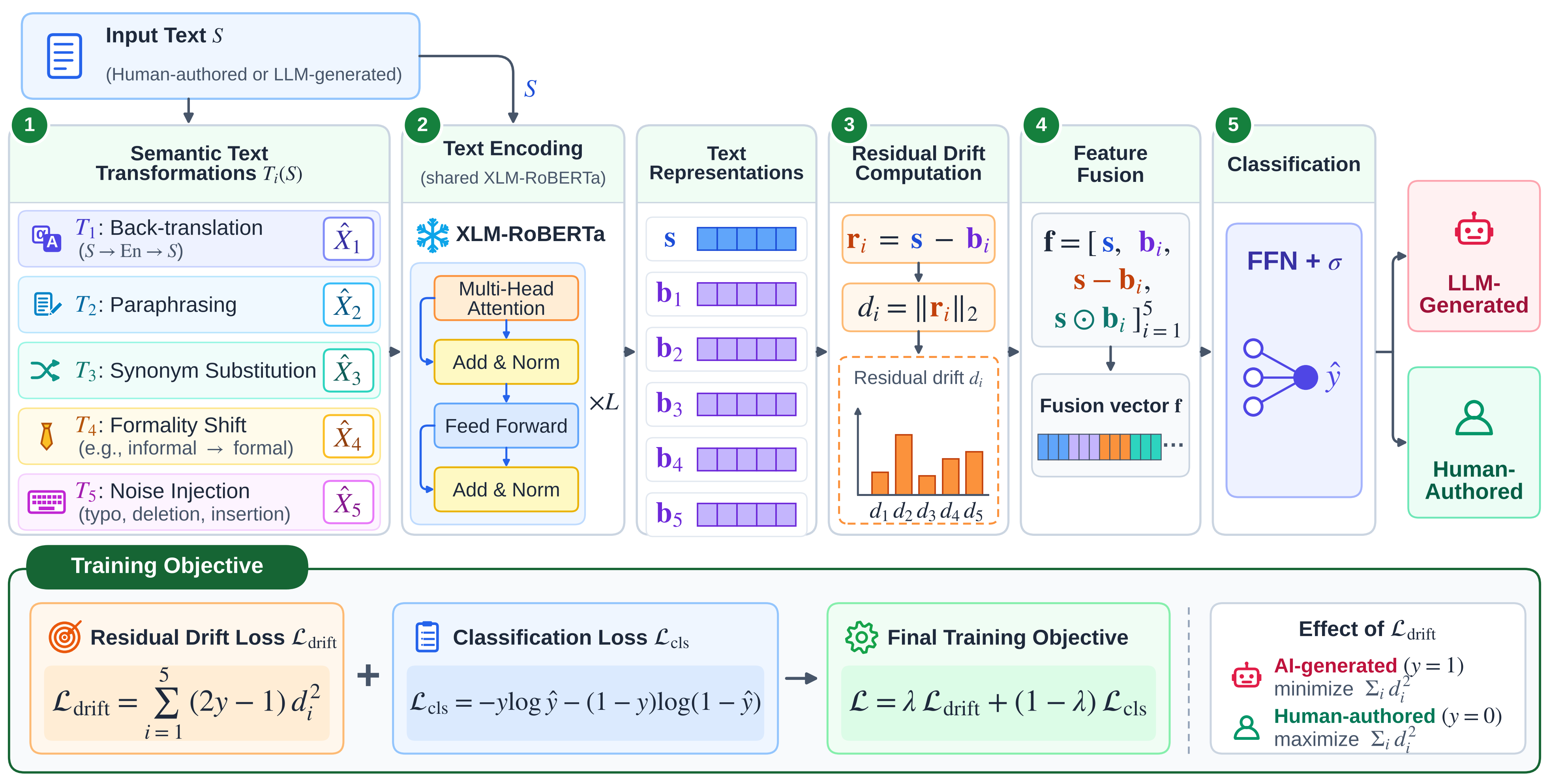} 
    \caption{ Overview of the proposed RED framework, where five LLM-generated semantics-preserving transformations and the original text are encoded using a shared XLM-RoBERTa encoder. Representation-level residual drift is computed between the original and transformed texts, followed by similarity–dissimilarity feature fusion and classification. A joint objective combines residual drift loss and classification loss to distinguish human-authored from LLM-generated text.}
    \label{fig:red}
    
\end{figure*}

\subsection{Residual Drift for AI-text Detection (RED)}
While the M-TRUST explicitly captures token-level divergences between the original and transformed text, RED focuses on capturing the loss of information in transformation by quantifying how much meaning or contextual nuance changes when the text is reconstructed through transformation operations.

We  first obtain and encoded representation $\mathbf{s}$, $\mathbf{b}_1$, $\mathbf{b}_2$, $\mathbf{b}_3$, $\mathbf{b}_4$ and  $\mathbf{b}_5$ of source text $\mathcal{S}$, $\acute{\mathcal{X}}_1$, $\acute{\mathcal{X}}_2$, $\acute{\mathcal{X}}_3$, $\acute{\mathcal{X}}_4$ and $\acute{\mathcal{X}}_5$ use XLM-RoBERTa encoder model.
\begin{equation}
\label{eq:encoder}
    \mathbf{s},\;  \mathbf{b}_1,\;  \mathbf{b}_2,\; \mathbf{b}_3,\; \mathbf{b}_4,\; \mathbf{b}_5 \; \text{=} f_{\theta}(\mathcal{S}), \; f_{\theta}(\acute{\mathcal{X}}_1 ), \; f_{\theta}(\acute{\mathcal{X}}_2), \; f_{\theta}(\acute{\mathcal{X}}_3), \; f_{\theta}(\acute{\mathcal{X}}_4), \;f_{\theta}(\acute{\mathcal{X}}_5)
\end{equation}
where $f_{\theta}(\cdot)$ denotes the XLM-RoBERTa encoder parameterized by $\theta$.
Figure~\ref{fig:red} presents the working of our proposed framework RED. 
Next, we estimate the residual drift vector $\mathbf{r}_i$ for each transformation as defined below.
\begin{equation}
\label{eq:rss}
    \mathbf{r}_i \; \text{=} \; \mathbf{s} - \mathbf{b}_i
\end{equation}
where $\mathbf{r}_i$ denotes the residual drift vector between the encoded representation $\mathbf{s}$ of original text  and the encoded representation  $\mathbf{b}_i$ of transformed text , as obtained from Equation~\ref{eq:encoder}. Subsequently we estimate the magnitude residual drift between the encoded representation $\mathbf{s}$ of the original text  and the encoded representation  $\mathbf{b}_i$ of transformed text
\begin{equation}
    d_i = \| \mathbf{r}_i \|_2,
\end{equation}
where $\| \cdot \|_2$ denotes the Euclidean norm.
Next, we define our Residual Drift Loss $\mathcal{L}_{\text{drift}}$ as a regularization objective that enforces low semantic drift for AI-generated text and high drift for human-authored text, thereby guiding the model to learn discriminative consistency patterns.
\begin{equation}
\label{eq:consistency_loss_di}
\mathcal{L}_{\text{drift}} =
\begin{cases}
\displaystyle \sum_{i=1}^{n} d_i^2, & \text{if AI-generated}, \\[8pt]
\displaystyle -\sum_{i=1}^{n} d_i^2, & \text{if Human-written.}
\end{cases}
\end{equation}
where $y \in \{0,1\}$ is the binary label and $d_i = \| \mathbf{s} - \mathbf{b}_i \|_2$ 
denotes the semantic drift magnitude for the $i^{th}$ transformation. 
The multiplicative term $(2y - 1)$ controls the direction of optimization 
based on the class label.
Similarly, the unified $\mathcal{L}_{\text{drift}}$ loss function is defined below.
\begin{equation}
\mathcal{L}_{\text{drift}} = \sum_{i=1}^{n} (2y - 1) \, d_i^2.
\end{equation}
For AI-generated text ($y = 1$), the coefficient $(2y - 1) = +1$, 
and the model minimizes $\sum_i d_i^2$, thereby enforcing semantic stability 
across transformations. This reflects the characteristic consistency of 
machine-generated text, which tends to preserve meaning even under 
back-translation or paraphrasing. In contrast, for human-authored text ($y = 0$), the coefficient 
$(2y - 1) = -1$, and the model maximizes $\sum_i d_i^2$, promoting 
greater semantic variability. This encourages the model to recognize 
that human writing often exhibits richer stylistic diversity, 
idiomatic flexibility, and contextual nuance that result in higher drift.

The intuition behind $\mathcal{L}_{\text{drift}}$ stems from the 
empirical observation that AI-generated text tends to remain 
semantically stable under meaning preserving transformations, 
whereas human-authored text exhibits greater variability. 
Hence, for AI-generated samples, the model is encouraged to 
minimize the residual drift (enforcing consistency), 
while for human-written text, it is encouraged to 
maximize the drift (capturing linguistic and stylistic flexibility). 
This bidirectional objective allows the model to learn the 
distinct semantic stability patterns characteristic of 
machine-generated and human-authored writing.\par 

Similarly, we also estimate the similarity and dissimilarity feature fusion vector $\mathbf{f}$ between the encoded representation $\mathbf{s}$ of the original text and the encoded representation  $\mathbf{b}_i$ of the transformed text by using Equation~\ref{eq:feature_fusion}.
\begin{align}
\label{eq:feature_fusion}
\mathbf{f} = & \;
\mathbf{s} \oplus \mathbf{b}_1 \oplus (\mathbf{s}-\mathbf{b}_1) \oplus (\mathbf{s} \odot \mathbf{b}_1)
\oplus \mathbf{s} \oplus \mathbf{b}_2 \oplus (\mathbf{s}-\mathbf{b}_2) \oplus (\mathbf{s} \odot \mathbf{b}_2) \notag\\
& \oplus \mathbf{s} \oplus \mathbf{b}_3 \oplus (\mathbf{s}-\mathbf{b}_3) \oplus (\mathbf{s} \odot \mathbf{b}_3)
\oplus \mathbf{s} \oplus \mathbf{b}_4 \oplus (\mathbf{s} - \mathbf{b}_4) \oplus (\mathbf{s} \odot \mathbf{b}_4) \notag\\
& \oplus \mathbf{s} \oplus \mathbf{b}_5 \oplus (\mathbf{s} - \mathbf{b}_5) \oplus (\mathbf{s} \odot \mathbf{b}_5)
\end{align}
Next, we feed the similarity and dissimilarity feature fusion vector $\mathbf{f}$ to a feedforward neural network.
\begin{equation}
\hat{y} = \sigma(\mathbf{W}_f \cdot \mathbf{f}  + \mathbf{b}_f)
\end{equation}
where $\mathbf{W}_f$ and $\mathbf{b}_f$ are learnable parameters. Following this, we define our classification loss function $\mathcal{C}_{cls}$ in Equation~\ref{eq:clasfication_loss}

\begin{equation}
\label{eq:clasfication_loss}
 \mathcal{C}_{cls} = -  y \log \hat{y} + (1-y) \log(1-\hat{y})
    \end{equation}

The final training objective of RED integrates two complementary components: the classification loss $\mathcal{L}_{\text{clf}}$ and the residual drift loss $\mathcal{L}_{\text{cons}}$. The overall learning objective is defined as a weighted combination, as shown in Equation~\ref{eq:final_loss}.

\begin{equation}
\label{eq:final_loss}
\mathcal{L} =  
\lambda \cdot \mathcal{L}_{\text{cons}}
+ (1-\lambda)\mathcal{L}_{\text{clf}}
\end{equation}
where $0 < \lambda < 1$ is  balances the classification and residual drift loss.


\begin{table}[t]
\centering
\caption{Implementation details and hyperparameter settings of the proposed M-TRUST and RED frameworks.}
\label{tab:hyperparams}
\begin{tabular}{ll}
\toprule
\rowcolor{blue!50!black}
\textcolor{white} {\textbf{Hyperparameter}} & \textcolor{white} {\textbf{Value}} \\
\midrule
Embedding Model & XLM-RoBERTa \\
\rowcolor[HTML]{D3D3D3} Maximum Sequence Length & 512 tokens \\ Number of Transformations & 5 \\
\rowcolor[HTML]{D3D3D3} Optimizer & AdamW \\
Batch Size & 16 \\
\rowcolor[HTML]{D3D3D3}Learning Rate & $2 \times 10^{-5}$ \\
Number of Epochs & 100 \\
\rowcolor[HTML]{D3D3D3}Output Activation & Sigmoid \\
Drift Metric & Euclidean Distance ($L_2$) \\
\rowcolor[HTML]{D3D3D3}Loss Weight ($\lambda$) & 0.1 \\
\bottomrule
\end{tabular}
\end{table}

\begin{table*}[t]
\centering
\caption{Performance comparison of the proposed and baseline models across multilingual datasets and genres. 
Here, {\it Acc} denotes overall accuracy, while {\it F1-H} and {\it F1-AI} represent the F1-scores for human-authored and AI-generated texts, respectively.
Bold values indicate the best performance in each category, and \textcolor{magenta}{Colour} values indicate the best performance.}
\label{tab:llm_detection_results_by_domain}
\resizebox{\textwidth}{!}{%
\begin{tabular}{ll|cccccccccccc|ccc|ccc|ccc|ccc}
\toprule
\rowcolor{blue!50!black}
\textcolor{white}{\textbf{Based}} & \textcolor{white} {\textbf{Dataset}}
& \multicolumn{12}{c}{\textcolor{white}{\textbf{SHAKTI}~\cite{agrahari2026trident}} }
& \multicolumn{3}{c}{\textcolor{white} {\textbf{$AG_{\text{hi}}$~\cite{kavathekar2024counter} (Hindi)}} }
& \multicolumn{3}{c}{\textcolor{white} {\textbf{M4~\cite{wang2024m4} (Arabic)}} }
& \multicolumn{3}{c}{\textcolor{white} {\textbf{KatFishNet~\cite{park2025katfishnet} (Korean)}} }
& \multicolumn{3}{c}{\textbf{\textcolor{white} {RAID~\cite{dugan2024raid}(English)}}} \\ 
\midrule
& \textbf{Model/Genres}
& \multicolumn{3}{c}{\textbf{QA}} 
& \multicolumn{3}{c}{\textbf{News}} 
& \multicolumn{3}{c}{\textbf{Wiki}} 
& \multicolumn{3}{c}{\textbf{Essay}}
& \multicolumn{3}{c}{\textbf{News}} 
& \multicolumn{3}{c}{\textbf{News}} 
& \multicolumn{3}{c}{\textbf{News}} 
& \multicolumn{3}{c}{\textbf{M4}} \\
& & \textbf{Acc} & \textbf{F1-H} & \textbf{F1-AI} 
& \textbf{Acc} & \textbf{F1-H} & \textbf{F1-AI} 
& \textbf{Acc} & \textbf{F1-H} & \textbf{F1-AI} 
& \textbf{Acc} & \textbf{F1-H} & \textbf{F1-AI}
& \textbf{Acc} & \textbf{F1-H} & \textbf{F1-AI} 
& \textbf{Acc} & \textbf{F1-H} & \textbf{F1-AI} 
& \textbf{Acc} & \textbf{F1-H} & \textbf{F1-AI} 
& \textbf{Acc} & \textbf{F1-H} & \textbf{F1-AI} \\ 
\midrule
\rowcolor[HTML]{D3D3D3} Baselines & Log-Likelihood~\cite{gehrmann2019gltr} & 0.29 & 0.28 & 0.31 & 0.29 & 0.26 & 0.32 & 0.32 & 0.30 & 0.34 & 0.27 & 0.25 & 0.29 & 0.30 & 0.28 & 0.31 & 0.45 & 0.36 & 0.47 & 0.40 & 0.37 & 0.46 & 0.54 & 0.44 & 0.65\\
& Rank~\cite{gehrmann2019gltr} & 0.43 & 0.41 & 0.44 & 0.45 & 0.44 & 0.46 & 0.44 & 0.43 & 0.45 & 0.42 & 0.41 & 0.43 & 0.45 & 0.44 & 0.47 & 0.46 & 0.42 & 0.49 & 0.44 & 0.41 & 0.47 & 0.66 & 0.53 & 0.68 \\
\rowcolor[HTML]{D3D3D3} & Log-Rank~\cite{gehrmann2019gltr} & 0.45 & 0.44 & 0.47 & 0.47 & 0.46 & 0.49 & 0.47 & 0.46 & 0.48 & 0.45 & 0.44 & 0.47 & 0.47 & 0.45 & 0.50 & 0.48 & 0.44 & 0.46 & 0.47 & 0.45 & 0.48 & 0.56 & 0.47 & 0.65  \\
& Entropy~\cite{gehrmann2019gltr} & 0.49 & 0.48 & 0.50 & 0.50 & 0.49 & 0.52 & 0.51 & 0.50 & 0.51 & 0.48 & 0.47 & 0.49 & 0.51 & 0.50 & 0.53 & 0.52 & 0.49 & 0.55 & 0.51 & 0.48 & 0.53  & 0.69 & 0.56 & 0.69  \\
\rowcolor[HTML]{D3D3D3} & NPR~\cite{su2023detectllm} & 0.64 & 0.65 & 0.64 & 0.60 & 0.59 & 0.61 & 0.63 & 0.70 & 0.52 & 0.59 & 0.57 & 0.61 & 0.61 & 0.59 & 0.64 & 0.69 & 0.61 & 0.73 & 0.73 & 0.70 & 0.76  & 0.80 & 0.79 & 0.85 \\
& DetectGPT~\cite{mitchell2023detectgpt} & 0.60 & 0.59 & 0.61 & 0.63 & 0.62 & 0.64& 0.61 & 0.60 & 0.62 & 0.63 & 0.61 & 0.66 & 0.64 & 0.62 & 0.66 & 0.69 & 0.65 & 0.73 & 0.68 & 0.65 & 0.69  & 0.81 & 0.77 & 0.87 \\
\rowcolor[HTML]{D3D3D3} &  Fast-DetectGPT~\cite{bao2024fast}& 0.78 & 0.80 & 0.77 & 0.77 & 0.74 & 0.80 & 0.79 & 0.77 & 0.81 & 0.77 & 0.75 & 0.80 & 0.82 & 0.81 & 0.84 & 0.74 & 0.71 & 0.77 & 0.73 & 0.70 & 0.76  & 0.82 & 0.75 & 0.88 \\
& J-Guard~\cite{kumarage2023j} & 0.79 & 0.81 & 0.77 & \textbf{0.82} & \textbf{0.83} & 0.81 & 0.78 & 0.77 & 0.79 & 0.81 & 0.80 & 0.83 & 0.80 & 0.81 & 0.78 & 0.78 & 0.74 & 0.81 & 0.77 & 0.73 & 0.79  & \textbf{0.90} & 0.81 & \textbf{0.90 }\\
\rowcolor[HTML]{D3D3D3} & RAIDAR~\cite{mao2024raidar} & 0.76 & 0.75 & 0.77 & 0.78 & 0.77 & 0.80 & 0.77 & 0.76 & 0.78 & 0.75 & 0.74 & 0.76 & 0.76 & 0.76 & 0.77  & \textbf{0.86} & 0.75 & \textbf{0.91} & 0.71 & 0.68 & 0.77  & 0.82 & 0.77 & 0.87 \\
& SimLLM~\cite{nguyen2024simllm} & 0.81 & 0.82 & 0.79 & 0.82 & 0.80 & \textbf{0.83} & \textbf{0.81} & 0.79 & \textbf{0.82} & 0.79 & 0.77 & 0.80 & \textbf{0.83} & \textbf{0.81} & \textcolor{blue}{0.84}
& 0.82 & 0.77 & 0.83 & 0.85 & 0.82  & 0.85 & 0.83 & 0.78  & 0.85  \\

\rowcolor[HTML]{D3D3D3} & LLaMA-L2R~\cite{hao-etal-2025-learning}& 0.79 & 0.77 & 0.80 & 0.79 & 0.75 & 0.82 & 0.79 & \textcolor{blue}{0.80} & 0.79 & 0.73 & 0.75 & 0.71 & 0.81 & 0.79 & 0.82 
& 0.82 & 0.79 & 0.84 & 0.80  & 0.74  & 0.83 & 0.84 & 0.79 & 0.85 \\

& BISCOPE~\cite{guo2024biscope} & \textbf{0.84} & \textcolor{blue}{0.86} & \textbf{0.83} & 0.81 & 0.79 & 0.82 & 0.80 & 0.78 & 0.81 & 0.77 & 0.75 & 0.78 & 0.82 & 0.80 & 0.83 & 0.82 & \textbf{0.77} & \textbf{0.86} & \textbf{0.85} & \textbf{0.76} & \textbf{0.88}  & 0.87 & \textbf{0.85} & \textbf{0.88}  \\
\rowcolor[HTML]{D3D3D3} & XLM-RoBERTa \cite{conneau2019cross}  &  0.82 & 0.64 & 0.72 & 0.81 & 0.67 & 0.65 & 0.83 & 0.75 & 0.74 & 0.81 & 0.66 & 0.74  & 0.84 & 0.70 & 0.73 & 0.83 & 0.69 & 0.71 & 0.82 & 0.68 & 0.70  & 0.79 & 0.44 & 0.81\\
\midrule
\rowcolor[HTML]{D3D3D3} \textbf{Proposed} & \textbf{M-TRUST  (mBART)} & 0.77 & 0.43 & 0.78 & 0.79 & 0.64 & 0.82 & 0.76 & 0.68 & 0.78 & 0.78 & 0.61 & 0.78  & 0.73 & 0.68 & 0.75 & 0.69 & 0.62 & 0.71 & 0.76 & 0.51 & 0.79  & 0.71 & 0.57 & 0.73 \\
& \textbf{M-TRUST  (GPT-3.5)} & 0.88 & 0.83 & 0.89 & \textbf{0.91} & \textbf{0.87} & \textbf{0.95} & \textbf{0.90} & \textbf{0.86} & \textbf{0.94} & \textbf{0.90} & \textbf{0.86} & \textbf{0.92} 
& \textbf{0.91} & \textbf{0.87} & \textbf{0.91} & \textbf{0.90} & \textbf{0.84} & \textbf{0.93} & 0.85 & 0.82 & 0.87  & \textbf{0.89} & \textbf{0.86} & 0.90 \\
\rowcolor[HTML]{D3D3D3} & \textbf{M-TRUST  (LLaMA-3.3)} & \textbf{0.89} & \textbf{0.86} & \textbf{0.89} & 0.87 & 0.84 & 0.88 & 0.86 & 0.83 & 0.87 & 0.87 & 0.84 & 0.89 
& 0.87 & 0.84 & 0.89 & 0.86 & 0.83 & 0.87 & \textbf{0.89} & \textbf{0.85} & \textbf{0.90}  & 0.86 & 0.84 & \textbf{0.90}   \\
& \textbf{M-TRUST  (Gemma-3)} & 0.87 & 0.81 & 0.89 & 0.85 & 0.82 & 0.85 & 0.84 & 0.81 & 0.86 & 0.85 & 0.82 & 0.85 
& 0.85 & 0.82 & 0.88 & 0.84 & 0.81 & 0.89 & 0.83 & 0.80 & 0.88  & 0.83 & 0.78 & 0.84\\
\rowcolor[HTML]{D3D3D3} & \textbf{M-TRUST  (Phi-4)} & 0.88 & 0.84 & 0.88 & 0.89 & 0.85 & 0.92 & 0.88 & 0.84 & 0.92 & 0.89 & 0.85 & 0.90 
& 0.87 & 0.85 & 0.87 & 0.88 & 0.84 & 0.88 & 0.87 & 0.83 & 0.91  & 0.84 & 0.77 & 0.88 \\ 
\cline{2-26}
& \textbf{RED (mBART)} & 0.78 & 0.63 & 0.79 & 0.81 & 0.64 & 0.82 & 0.71 & 0.62 & 0.71 & 0.77 & 0.61 & 0.80  & 0.78 & 0.63 & 0.79 & 0.70 & 0.52 & 0.85 & 0.76 & 0.71 & 0.79  & 0.81 & 0.79 & 0.83   \\
\rowcolor[HTML]{D3D3D3} & \textbf{RED (GPT-3.5)} & \textcolor{magenta}{\textbf{0.94}} & \textcolor{magenta}{\textbf{0.94}} & \textcolor{magenta}{\textbf{0.92}} & \textcolor{magenta}{\textbf{0.96}} & \textcolor{magenta}{\textbf{0.95}} & 0.98 & \textcolor{magenta}{\textbf{0.97}} & \textcolor{magenta}{\textbf{0.94}} & \textcolor{magenta}{\textbf{0.97}} & \textcolor{magenta}{\textbf{0.98}} & \textcolor{magenta}{\textbf{0.95}} & \textcolor{magenta}{\textbf{0.98}} 
& 0.88 & 0.85 & 0.89 & 0.87 & 0.84 & 0.92 & 0.91 & 0.83 & \textcolor{magenta}{\textbf{0.91}}  & 0.88 &\textcolor{magenta}{\textbf{ 0.87}} & \textcolor{magenta}{\textbf{0.91}}\\
& \textbf{RED (LLaMA-3.3)} & 0.90 & 0.87 & 0.91 & \textbf{0.96} & 0.88 & \textcolor{magenta}{\textbf{0.96}} & 0.91 & 0.87 & 0.92 & 0.91 & 0.87 & 0.91 
& \textcolor{magenta}{\textbf{0.92}} & \textcolor{magenta}{\textbf{0.88}} & \textbf{0.94} & \textcolor{magenta}{\textbf{0.91}} & \textcolor{magenta}{\textbf{0.87}} & 0.93 & \textcolor{magenta}{\textbf{0.92}} & \textcolor{magenta}{\textbf{0.86}} & \textcolor{magenta}{\textbf{0.94}} & \textcolor{magenta}{\textbf{0.91}} & 0.88 & \textbf{0.91} \\
\rowcolor[HTML]{D3D3D3} & \textbf{RED (Gemma-3)} & 0.85 & 0.82 & 0.86 & 0.86 & 0.83 & 0.87 & 0.85 & 0.82 & 0.86 & 0.86 & 0.83 & 0.86 
& 0.86 & 0.83 & 0.88 & 0.85 & 0.82 & 0.86 & 0.84 & 0.81 & 0.89  & 0.88 & 0.85 & 0.88   \\
& \textbf{RED (Phi-4)} & 0.89 & 0.87 & 0.91 & 0.90 & 0.86 & 0.94 & 0.89 & 0.85 & 0.89 & 0.90 & 0.86 & 0.94 
& 0.91 & 0.87 & \textcolor{magenta}{\textbf{0.92}} & 0.90 & 0.86 & \textcolor{magenta}{\textbf{0.90}} & 0.89 & 0.85 & 0.89  & 0.88 & 0.88 & \textcolor{magenta}{\textbf{0.94}} \\ 
\bottomrule
\end{tabular}%
}
\end{table*}

\begin{table}[t]
\centering
\caption{Performance of the proposed \textit{RED (LLaMA 3.3)}
framework under individual linguistic transformations across five
datasets: SHAKTI, $AG_{\mathrm{hi}}$ (Hindi)~\cite{kavathekar2024counter},
M4 (Arabic)~\cite{wang2024m4}, KatFishNet (Korean),
and M4 (English)~\cite{wang2024m4}.}
\label{tab:transformations}
\setlength{\tabcolsep}{3pt}
\begin{adjustbox}{max width=\textwidth}
\begin{tabular}{l|cc|cc|cc|cc|cc}
\toprule
\rowcolor{blue!50!black}
\textcolor{white} {\textbf{Transformations}}
& \multicolumn{2}{c|}{\textcolor{white} {\textbf{SHAKTI}}}
& \multicolumn{2}{c|}{\textcolor{white} {\textbf{$AG_{\mathrm{hi}}$}}}
& \multicolumn{2}{c|}{\textcolor{white} {\textbf{Arabic}}}
& \multicolumn{2}{c|}{\textcolor{white} {\textbf{Korean}}}
& \multicolumn{2}{c}{\textcolor{white} {\textbf{English}}} \\
& \textbf{Acc} & \textbf{F1}
& \textbf{Acc} & \textbf{F1}
& \textbf{Acc} & \textbf{F1}
& \textbf{Acc} & \textbf{F1}
& \textbf{Acc} & \textbf{F1} \\
\midrule

\rowcolor[HTML]{D3D3D3} Back-Translation
& \textcolor{magenta}{\textbf{0.97}}
& \textcolor{magenta}{\textbf{0.95}}
& 0.93 & 0.91
& \textbf{0.91} & \textbf{0.88}
& \textbf{0.89} & \textbf{0.86}
& -- & -- \\

Paraphrasing
& 0.95 & 0.94
& \textbf{0.95} & \textbf{0.93}
& 0.89 & 0.86
& 0.87 & 0.84
& \textbf{0.96} & \textbf{0.94} \\

\rowcolor[HTML]{D3D3D3} Lexical Substitution
& 0.88 & 0.86
& 0.86 & 0.84
& 0.81 & 0.78
& 0.78 & 0.75
& 0.89 & 0.88 \\

Formality Shifts
& 0.91 & 0.90
& 0.89 & 0.87
& 0.83 & 0.80
& 0.82 & 0.79
& 0.90 & 0.89 \\

\rowcolor[HTML]{D3D3D3} Noise Injection
& 0.80 & 0.76
& 0.77 & 0.72
& 0.71 & 0.67
& 0.69 & 0.64
& 0.78 & 0.74 \\

\bottomrule
\end{tabular}
\end{adjustbox}
\end{table}

\begin{table*}[t]
\centering
\caption{Performance comparison of the proposed M-TRUST  and RED across multilingual subsets of the \textbf{M4} dataset~\cite{wang2024m4}.}
\label{tab:multilingual_m4_results}
\small
\setlength{\tabcolsep}{3pt}

\begin{adjustbox}{max width=\textwidth}
\begin{tabular}{lcccccccccccccccc}
\toprule
\rowcolor{blue!50!black}
\textcolor{white} { \textbf{Model / Dataset}}
& \multicolumn{2}{c|}{\textcolor{white}{\textbf{English}}}
& \multicolumn{2}{c|}{\textcolor{white} {\textbf{Urdu}}}
& \multicolumn{2}{c|}{\textcolor{white} {\textbf{Russian}}}
& \multicolumn{2}{c|}{\textcolor{white} {\textbf{Bulgarian}}}
& \multicolumn{2}{c|}{\textcolor{white} {\textbf{Chinese}}}
& \multicolumn{2}{c|}{\textcolor{white} {\textbf{German}}}
& \multicolumn{2}{c|}{\textcolor{white} {\textbf{Indonesian}}}
& \multicolumn{2}{c}{\textcolor{white} {\textbf{Italian}}}\\

& \textbf{Acc} & \textbf{F1}
& \textbf{Acc} & \textbf{F1}
& \textbf{Acc} & \textbf{F1}
& \textbf{Acc} & \textbf{F1}
& \textbf{Acc} & \textbf{F1}
& \textbf{Acc} & \textbf{F1}
& \textbf{Acc} & \textbf{F1}
& \textbf{Acc} & \textbf{F1}\\

\midrule

\rowcolor[HTML]{D3D3D3} M-TRUST  (GPT-3.5)
& 0.90 & 0.87
& 0.88 & 0.85
& 0.89 & 0.86
& 0.87 & 0.84
& 0.86 & 0.83
& 0.82 & 0.79
& 0.83 & 0.80
& 0.85 & 0.82 \\

M-TRUST  (LLaMA-3.3)
& 0.89 & 0.86
& 0.87 & 0.84
& 0.88 & 0.85
& 0.86 & 0.83
& 0.85 & 0.82
& \textbf{0.84} & \textbf{0.81}
& 0.82 & 0.79
& 0.83 & 0.80 \\

\rowcolor[HTML]{D3D3D3} M-TRUST  (Gemma-3)
& 0.88 & 0.85
& \textbf{0.89} & \textbf{0.86}
& 0.87 & 0.84
& 0.85 & 0.82
& 0.84 & 0.81
& 0.83 & 0.80
& \textbf{0.85} & \textbf{0.82}
& 0.82 & 0.79 \\

M-TRUST  (Phi-4)
& 0.87 & 0.84
& 0.85 & 0.82
& 0.86 & 0.83
& 0.84 & 0.81
& 0.83 & 0.80
& 0.81 & 0.78
& 0.81 & 0.78
& 0.84 & 0.81 \\

\midrule

\rowcolor[HTML]{D3D3D3} RED (GPT-3.5)
& \textbf{0.95} & \textbf{0.93}
& 0.94 & 0.91
& \textbf{0.95} & \textbf{0.92}
& 0.93 & 0.90
& \textbf{0.94} & \textbf{0.91}
& 0.90 & 0.87
& 0.91 & 0.88
& \textbf{0.94} & \textbf{0.91} \\

RED (LLaMA-3.3)
& 0.94 & 0.92
& \textbf{0.95} & \textbf{0.92}
& 0.94 & 0.91
& \textbf{0.94} & \textbf{0.91}
& 0.92 & 0.89
& \textbf{0.92} & \textbf{0.89}
& 0.90 & 0.87
& 0.92 & 0.89 \\

\rowcolor[HTML]{D3D3D3} RED (Gemma-3)
& 0.93 & 0.90
& 0.92 & 0.89
& 0.92 & 0.89
& 0.91 & 0.88
& 0.93 & 0.90
& 0.91 & 0.88
& \textbf{0.93} & \textbf{0.90}
& 0.91 & 0.88 \\

RED (Phi-4)
& 0.92 & 0.89
& 0.90 & 0.87
& 0.91 & 0.88
& 0.92 & 0.89
& 0.90 & 0.87
& 0.89 & 0.86
& 0.89 & 0.86
& 0.92 & 0.89 \\

\bottomrule
\end{tabular}
\end{adjustbox}
\end{table*}

\section{Experimental Setup}
To evaluate and compare the performance of our proposed methods, we consider several state-of-the-art approaches from the literature as baselines, including \textit{Log-Likelihood}, \textit{Rank}, \textit{Log-Rank}, and \textit{Entropy}~\cite{gehrmann2019gltr}, \textit{NPR}~\cite{su2023detectllm}, \textit{DetectGPT}~\cite{mitchell2023detectgpt}, \textit{Fast-DetectGPT}~\cite{bao2024fast}, and \textit{J-Guard}~\cite{kumarage2023j}. Furthermore, we consider three transformation-based detection methods as baselines: RAIDAR~\cite{mao2024raidar} and LLaMA-L2R~\cite{hao-etal-2025-learning}, \textit{BISCOPE}~\cite{guo2024biscope}, which exploit LLM-based rewriting behaviour, and SimLLM~\cite{nguyen2024simllm}, which leverages similarity between original and LLM-regenerated texts. Additionally, we fine-tune \textit{XLM-RoBERTa}~\cite{conneau2019cross} as a baseline model. We independently employ \textbf{GPT-3.5}, \textbf{LLaMA-3.3}, \textbf{Gemma-3}, and \textbf{Phi-4} to apply the transformations $T_i(\mathcal{S})$ to the original source text $\mathcal{S}$, as defined in Equation~\ref{Eq:transformation} Additionally, we employ \textbf{mBART}\footnote{\url{https://huggingface.co/docs/transformers/en/model_doc/mbart}} as an alternative to LLMs for performing one-round back-translation on the original source text $\mathcal{S}$, as defined in Equation~\ref{Eq:transformation} for both M-TRUST and RED.

We evaluate the performance of our proposed methods and baselines on a diverse set of datasets across multiple languages and domains to assess their robustness and generalizability for LLM-generated text detection. Specifically, we use the Hindi datasets $AG_{\mathrm{hi}}$~\cite{kavathekar2024counter} and \textit{SHAKTI}~\cite{agrahari2026trident}, the Arabic dataset M4~\cite{wang2024m4}, the Korean dataset KatFishNet~\cite{park2025katfishnet}, and the English dataset RAID~\cite{dugan2024raid}. The \textit{SHAKTI} dataset comprises four genres: Question Answering (QA), News, Wikipedia, and Essay, enabling genre-wise evaluation within Hindi. 

Furthermore, to evaluate our methods across additional multilingual languages, we employ the \textbf{M4}~\cite{wang2024m4} dataset, a \textbf{m}ulti-generator, \textbf{m}ulti-domain, and \textbf{m}ultilingual benchmark for \textbf{m}achine-generated text detection. M4 comprises language-specific datasets covering Arabic, Bulgarian, Chinese, English, Indonesian, Russian, and Urdu. We use the predefined training, validation, and test splits from the original datasets when available; otherwise, we randomly split the data into training, validation, and test sets using a 70:10:20 ratio. We evaluate model performance using accuracy and class-specific F1-scores for the human-written (F1-H) and AI-generated (F1-AI) classes. Table \ref{tab:hyperparams} presents further details of the hyperparameters and our experimental setup.

\section{Results and Discussion}
Table~\ref{tab:llm_detection_results_by_domain} presents the performance comparison between the proposed and baseline models on
$AG_{\mathrm{hi}}$~\cite{kavathekar2024counter} (Hindi), M4~\cite{wang2024m4} (Arabic), KatFishNet~\cite{park2025katfishnet} (Korean) and  RAID~\cite{dugan2024raid} (English) datasets across four languages. From Table~\ref{tab:llm_detection_results_by_domain}, it is evident that our proposed methods \textit{M-TRUST} and \textit{RED} outperform the baseline models across five  datasets spanning  four languages and genres. Among the baseline methods, \textit{BISCOPE}, \textit{RAIDAR}, and \textit{J-Guard} demonstrate competitive performance but exhibit limited
generalizability across low-resource languages and multilingual settings. Next, Table~\ref{tab:llm_detection_results_by_domain} demonstrate that our proposed methods, M-TRUST  and RED, outperformed 
transformation-based baselines, including RAIDAR~\cite{mao2024raidar}, SimLLM~\cite{nguyen2024simllm}, and LLaMA-L2R~\cite{hao-etal-2025-learning}, across all evaluated datasets and genres.  From this observation, we conclude that leveraging transformation-induced differences in contextual representations yields more discriminative signals than relying solely on surface-level modifications or text similarity. Also, RAIDAR~\cite{mao2024raidar} and SimLLM~\cite{nguyen2024simllm} primarily rely on rewriting or text regeneration to identify LLM-generated content, and the effectiveness of such transformations may vary across languages, particularly in low-resource settings where the linguistic capabilities of LLMs can be limited. In contrast, our proposed methods leverage multiple meaning-preserving transformations to capture diverse linguistic variations, demonstrating improved detection performance across the evaluated languages. Both the baseline (Fine-tuning XLM-RoBERTa) and the proposed M-TRUST  method employ the same XLM-RoBERTa backbone, with the baseline operating directly on the original text without transformations. As shown in Table~\ref{tab:llm_detection_results_by_domain}, M-TRUST   consistently outperform the fine-tuned XLM-RoBERTa baseline across datasets and languages.  From these observations, we conclude that the performance gains of the M-TRUST stem from the transformation-aware representation-learning framework rather than solely from fine-tuning the underlying pretrained encoder. Among the proposed methods, RED consistently outperform M-TRUST , suggesting that residual representation differences provide effective discriminative signals for distinguishing human-authored from LLM-generated text. Table~\ref{tab:llm_detection_results_by_domain} demonstrates that M-TRUST  and RED generally achieve superior detection performance when employing diverse text transformations using GPT-3.5, LLaMA-3.3, Gemma-3, or Phi-4 compared with mBART-based round-trip translation.

\textbf{Impact of Semantic Transformation Strategies on the  Performance of RED and M-TRUST  :} 
To study the influence of semantic transformation $T_i(\mathcal{S})$ and how the choice of transformation affects the performance of the RED model, we conduct an empirical study by applying one semantic transformation at a time in RED instead of combining all transformations together. 
From Table~\ref{tab:transformations}, it is evident that although the performance of the RED model remains consistent across different semantic transformations, some transformations are slightly more effective in specific languages. The one-round back-translation transformation demonstrates greater effectiveness for Hindi, Korean, and Arabic, whereas the \textit{rephrasing} transformation proves more effective for detecting AI-generated text in English. From such observations, we conclude that the linguistic and structural characteristics of the target language influence the effectiveness of a semantic transformation in AI-generated text. However, as evident from Table~\ref{tab:llm_detection_results_by_domain}, combining all the semantic transformations, the performance of both \textbf{RED} and \textbf{M-TRUST } becomes stable and consistent across languages.

\textbf{Multilingual Robustness and Generalizability of RED and M-TRUST :}
To evaluate the multilingual robustness and generalizability of RED and M-TRUST , we conduct experiments across diverse languages. Specifically, we evaluate their performance on datasets covering \textit{Hindi}, \textit{Arabic}, \textit{Korean}, and \textit{English}, as presented in Table~\ref{tab:llm_detection_results_by_domain}, and on five additional language-specific datasets from M4~\cite{wang2024m4}, covering \textit{English}, \textit{Urdu}, \textit{Russian}, \textit{Bulgarian}, \textit{Chinese}, \textit{German}, \textit{Indonesian} and \textit{Italian}  as reported in Table~\ref{tab:multilingual_m4_results}. The primary motivation for investigating the performance of RED and M-TRUST on datasets from multiple languages is to verify whether our proposed semantic transformation-based methods, M-TRUST and RED, efficiently distinguish human-authored from AI-generated text across linguistically diverse languages. Table~\ref{tab:multilingual_m4_results} shows that the performance of our proposed methods is consistent across five languages, such as \textit{English}, \textit{Urdu}, \textit{Russian}, \textit{Bulgarian}, and \textit{Chinese} in the M4~\cite{wang2024m4} dataset. Considering the robust and consistent performance of RED and M-TRUST  in Table~\ref{tab:llm_detection_results_by_domain} and ~\ref{tab:multilingual_m4_results}, we conclude that semantic transformation, consistency, and residual drift-based cues generalise effectively in multilingual contexts. From these observations, we conclude that our proposed methods, RED and M-TRUST, are effective at detecting AI-generated text across linguistically diverse languages, including low-resource languages, by adapting the transformer-based encoder to the source languages.

\section{Generalization Across Transformations and Models}
\label{sec:generalization}
To assess the robustness and generalizability of M-TRUST and RED, we conduct experiments under two evaluation settings: (i) Cross-Transformation Generalisation and (ii) Generalisation to Unseen Text Generators and Transformation Models.

\textbf{Cross-Transformation Generalisation :}
We evaluate the ability of M-TRUST and RED to generalise across a range of semantics-preserving text transformations. Here, we consider five transformation strategies: paraphrasing, back-translation, lexical substitution, formality shifts, and noise injection. To study the cross-transformation generalizability of RED and M-TRUST, train each method using a single transformation strategy at a time, and evaluate their performance on both seen and unseen transformations. Table~\ref{tab:cross_transform_f1} presents the cross-transformation performance of M-TRUST and RED models. The diagonal entries correspond to in-domain evaluation; off-diagonal entries represent cross-transformation transfer. Table~\ref{tab:cross_transform_f1} reveals that both RED and M-TRUST exhibit strong performance on seen transformations, while their generalisation to unseen transformations varies considerably. Paraphrasing and back-translation demonstrate strong bidirectional transferability, suggesting that these transformations capture shared detection cues. In contrast, lexical substitution, formality shifts, and noise injection exhibit more transformation-specific behaviour, indicating that the learned representations are not entirely transformation-invariant. Furthermore, M-TRUST demonstrates stronger overall generalisation than RED, suggesting that its learned representations capture more transferable discriminative characteristics. Interestingly, the asymmetric transfer patterns across certain transformation pairs indicate that generalisation depends not only on the transformation encountered during testing but also on the transformation used during training. From such observation we conclude that M-TRUST and RED learn partially transferable detection signals, while their effectiveness remains influenced by the nature and compatibility of the training and testing transformations. Furthermore, the performance of M-TRUST and RED Table~\ref{tab:llm_detection_results_by_domain} using all five transformations is superior to that of RED and M-TRUST using any subset of transformations, suggesting that combining diverse transformation strategies provides complementary detection signals and enhances overall robustness.

\begin{table*}[t]
\centering
\caption{Cross-transformation generalisation on Hindi datasets \textit{SHAKTI}~\cite{agrahari2026trident}. Rows specify the training transformation, and columns specify the test transformation.}
\label{tab:cross_transform_f1}
\small
\setlength{\tabcolsep}{3pt}
\begin{adjustbox}{max width=\textwidth}
\begin{tabular}{llcccccccccc}
\toprule

\rowcolor{blue!50!black}
\textcolor{white} {Method} & \textcolor{white} {transformation} &\multicolumn{2}{c}{\textcolor{white}{Paraphrasing}}
& \multicolumn{2}{c}{\textcolor{white} {Back-translation}}
& \multicolumn{2}{c}{\textcolor{white}{Lexical substitution}}
& \multicolumn{2}{c}{\textcolor{white}{Formality shifts}}
& \multicolumn{2}{c}{\textcolor{white}{Noise injection}} \\
\cmidrule(lr){3-4}
\cmidrule(lr){5-6}
\cmidrule(lr){7-8}
\cmidrule(lr){9-10}
\cmidrule(lr){11-12}
\rowcolor[HTML]{D3D3D3}  & 
& Acc. & Micro-F1
& Acc. & Micro-F1
& Acc. & Micro-F1
& Acc. & Micro-F1
& Acc. & Micro-F1 \\
\midrule

\multirow{5}{*}{M-TRUST}
& Paraphrasing
& 0.980 & 0.980 & 0.969 & 0.969 & 0.891 & 0.891 & 0.872 & 0.872 & 0.841 & 0.841 \\

\rowcolor[HTML]{D3D3D3} & Back-translation
& 0.953 & 0.953 & 0.964 & 0.964 & 0.887 & 0.887 & 0.861 & 0.861 & 0.832 & 0.832 \\

& Lexical substitution
& 0.882 & 0.882 & 0.891 & 0.891 & 0.971 & 0.971 & 0.873 & 0.873 & 0.851 & 0.851 \\

\rowcolor[HTML]{D3D3D3}&  Formality shifts
& 0.891 & 0.891 & 0.881 & 0.881 & 0.897 & 0.897 & 0.984 & 0.984 & 0.862 & 0.862 \\

& Noise injection
& 0.841 & 0.841 & 0.852 & 0.852 & 0.832 & 0.832 & 0.861 & 0.861 & 0.968 & 0.968 \\
\midrule

\multirow{5}{*}{RED}
 & Paraphrasing
& 0.951 & 0.951 & 0.949 & 0.949 & 0.861 & 0.861 & 0.841 & 0.841 & 0.812 & 0.812 \\

\rowcolor[HTML]{D3D3D3} & Back-translation
& 0.957 & 0.957 & 0.956 & 0.956 & 0.852 & 0.852 & 0.832 & 0.832 & 0.801 & 0.801 \\

 &  Lexical substitution
& 0.851 & 0.851 & 0.862 & 0.862 & 0.961 & 0.961 & 0.841 & 0.841 & 0.821 & 0.821 \\

\rowcolor[HTML]{D3D3D3} & Formality shifts
& 0.861 & 0.861 & 0.851 & 0.851 & 0.872 & 0.872 & 0.958 & 0.958 & 0.812 & 0.812 \\

 &  Noise injection
& 0.812 & 0.812 & 0.821 & 0.821 & 0.801 & 0.801 & 0.823 & 0.823 & 0.947 & 0.947 \\

\bottomrule
\end{tabular}
\end{adjustbox}
\end{table*}

\textbf{Generalization to Unseen Text Generators and Transformation Models :}
To investigate whether the M-TRUST and RED learn generalizable discriminative patterns rather than generator- or transformation-model-specific characteristics, we evaluate their performance under four experimental settings: (i) seen text generators and transformation models (SG-ST), (ii) unseen text generators and seen transformation model (UG-ST), (iii) seen generator and unseen transformation models(SG-UT), and (iv) jointly unseen text generators and transformation models (UG-UT). Table~\ref{tab:unseen_gen_tf} presents the performance of M-TRUST and RED under four experimental settings to assess their generalizability across seen and unseen text generators and transformation models. Under the in-distribution setting (SG--ST), both M-TRUST and RED achieve strong detection performance, with RED marginally outperforming M-TRUST. Under the in-distribution setting (SG--ST), both M-TRUST and RED achieve strong detection performance, with RED marginally outperforming M-TRUST.  When exposed to unseen text generators (UG--ST), both models exhibit slight performance degradation, indicating strong cross-generator generalisation. However, under unseen transformation models (SG--UT), both models experience a noticeable performance decline, with M-TRUST exhibiting greater robustness than RED. When both the text generator and transformation model are unseen (UG--UT), the performance declines further. From such observations, we can conclude that both M-TRUST and RED generalise well to unseen text generators, suggesting that the discriminative signals captured through semantic transformations are largely transferable across text generators. However, both methods exhibit greater sensitivity to changes in the transformation model. While RED performs better when the transformation model remains unchanged, M-TRUST exhibits greater robustness to unseen transformation models. In summary , these findings highlight the potential of semantic transformation-based approaches for generator-agnostic AI-generated text detection while emphasising the need for improved generalisation across transformation models.

\begin{table*}[t]
\centering
\caption{Generalization to unseen text generators and transformation models on the Hindi SHAKTI dataset~\cite{agrahari2026trident}.}
\label{tab:unseen_gen_tf}
\footnotesize
\setlength{\tabcolsep}{4pt}
\renewcommand{\arraystretch}{1.2}

\begin{adjustbox}{width=\textwidth}
\begin{tabular}{@{}lccccc lcc@{}}
\toprule
\rowcolor{blue!50!black}
\textcolor{white}{\textbf{Evaluation Setting}} &
\multicolumn{2}{c}{\textcolor{white}{\textbf{Text Generator}}} &
\multicolumn{2}{c}{\textcolor{white}{\textbf{Transformation Model}}} &
\textcolor{white}{\textbf{Setting}} &
\textcolor{white}{\textbf{Method}} &
\textcolor{white}{\textbf{Accuracy}} &
\textcolor{white}{\textbf{Micro-F1}} \\

\cmidrule(lr){2-3}
\cmidrule(lr){4-5}
& \textbf{Training} & \textbf{Testing}
& \textbf{Training} & \textbf{Testing}
& & & & \\
\midrule

\rowcolor[HTML]{D3D3D3}
\multicolumn{9}{l}{\textbf{A. In-distribution (SG--ST)}} \\

& GPT-3.5 & GPT-3.5 & Qwen2.5-7B & Qwen2.5-7B & SG--ST
& M-TRUST & 0.984 & 0.985 \\

& GPT-3.5 & GPT-3.5 & Qwen2.5-7B & Qwen2.5-7B & SG--ST
& RED & \textbf{0.991} & \textbf{0.992} \\

\midrule

\rowcolor[HTML]{D3D3D3}
\multicolumn{9}{l}{\textbf{B. Unseen Generator (UG--ST)}} \\

& GPT-3.5 & Unseen & Qwen2.5-7B & Qwen2.5-7B & UG--ST
& M-TRUST & 0.982 & 0.984 \\

& GPT-3.5 & Unseen & Qwen2.5-7B & Qwen2.5-7B & UG--ST
& RED & \textbf{0.989} & \textbf{0.991} \\

\midrule

\rowcolor[HTML]{D3D3D3}
\multicolumn{9}{l}{\textbf{C. Unseen Transformation (SG--UT)}} \\

& GPT-3.5 & GPT-3.5 & Qwen2.5-7B & Llama-3.1-8B & SG--UT
& M-TRUST & \textbf{0.891} & \textbf{0.895} \\

& GPT-3.5 & GPT-3.5 & Qwen2.5-7B & Llama-3.1-8B & SG--UT
& RED & 0.872 & 0.878 \\

\midrule

\rowcolor[HTML]{D3D3D3}
\multicolumn{9}{l}{\textbf{D. Both Unseen (UG--UT)}} \\

& GPT-3.5 & Unseen & Qwen2.5-7B & Llama-3.1-8B & UG--UT
& M-TRUST & \textbf{0.862} & \textbf{0.868} \\

& GPT-3.5 & Unseen & Qwen2.5-7B & Llama-3.1-8B & UG--UT
& RED & 0.841 & 0.848 \\

\bottomrule
\end{tabular}
\end{adjustbox}

\vspace{2pt}
\begin{minipage}{\textwidth}
\footnotesize
\textbf{Note:} Unseen text generators include GPT-4, Bard, and Gemma-7B. SG/UG denote seen/unseen text generators, while ST/UT denote seen/unseen transformation models. Here, seen indicates that the same model is used for text generation or transformation during both training and testing, whereas unseen indicates that different models are used during training and testing.
\end{minipage}
\end{table*}

\begin{figure}[t]
\centering
\includegraphics[width=0.85\linewidth]{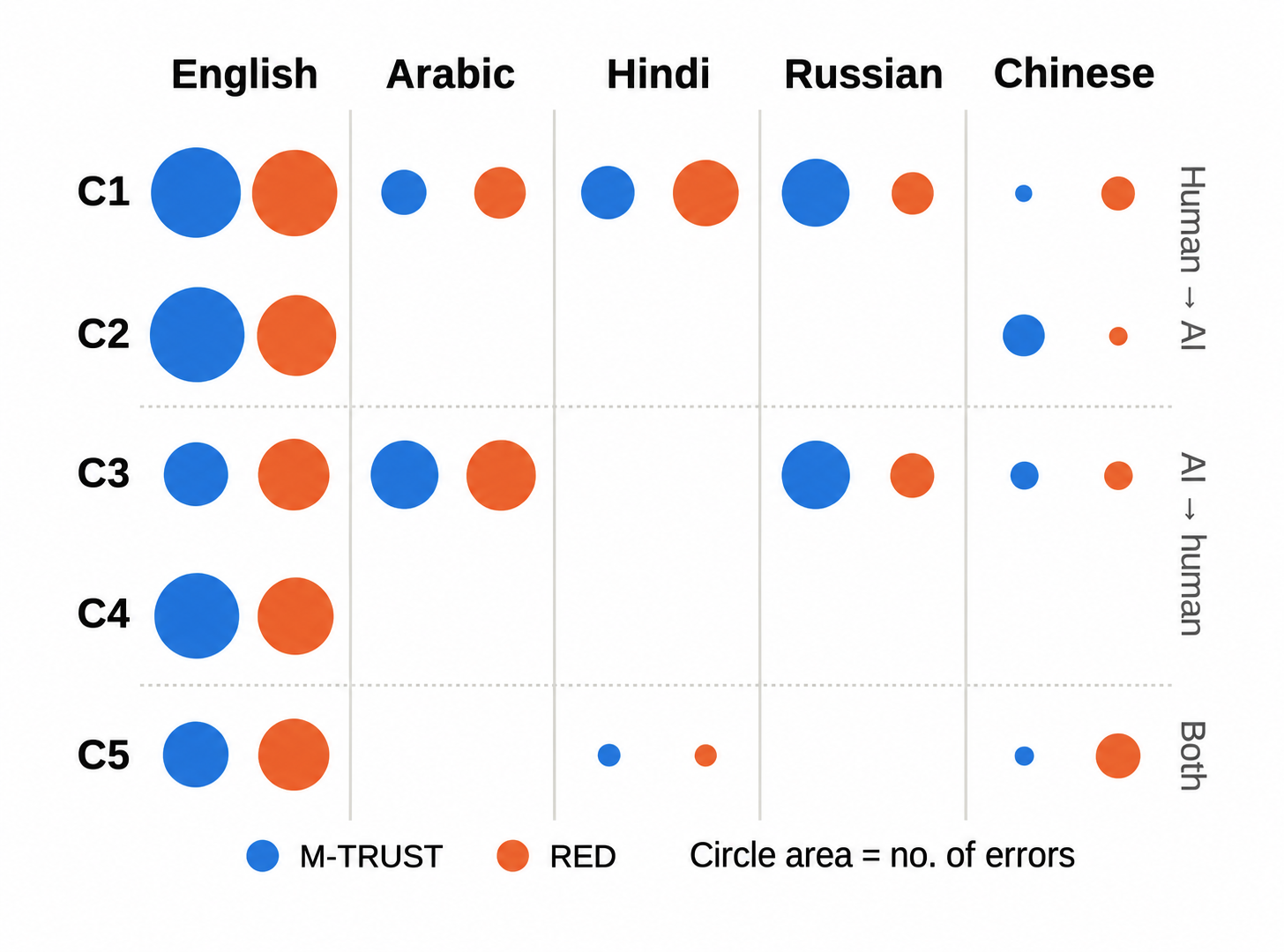}
\caption{Error categories by detector and language. Circle area is proportional to the number of misclassified samples; empty cells indicate no errors.}
\label{fig:error-map}
\end{figure}
\section{Error Analysis}
\label{sec:error-analysis}
We conduct an error analysis of samples misclassified by RED and M-TRUST. We conduct an error analysis by manually investigating the samples across all five languages. Figure~\ref{fig:error-map} presents our error analysis. The following key observations can be drawn from Figure~\ref{fig:error-map}. \textbf{C1: Formal or template-like human text :} (human $\rightarrow$ AI)  In case the source text is factual statements or quotation-based texts, sometimes misclassified as AI-generated because of their structured and predictable writing styles. \textbf{C2: Polished casual human text} (human $\rightarrow$ AI): Informal human-written text is sometimes misclassified as AI-generated because its clear and well-organised writing style lacks the errors and irregularities often associated with human writing. \textbf{C3: AI text in the human data style} (AI $\rightarrow$ human). When the generator matches the register and domain of the human data, no stylistic cue remains; even factual errors in such text go unnoticed. \textbf{C4: AI text imitating human casualness} (AI $\rightarrow$ human). A casual tone, a personal voice or deliberate imperfections make generated text pass as human.
\textbf{C5: Short or low-content text} (AI $\leftrightarrow$ human). Texts that are short in length provide too little evidence for a reliable decision. C1 and C3 are the most common error categories for both RED and M-TRUST, suggesting that these errors are more likely related to text characteristics than to differences between the two models. M-TRUST is more prone to C2, while RED struggles more with C5.
Error profiles also differ by language: C3 dominates Arabic and Russian, C1 dominates Hindi, and English exhibits all five error categories. In summary, C1, C2, and C4 highlight the challenges of distinguishing human-authored and AI-generated texts with overlapping stylistic characteristics. These findings suggest that similarities in writing style can contribute to misclassification, even when contextual and semantic features are considered.

 \section{Conclusion}
This study proposes two novel approaches, \textit{M-TRUST} and \textit{RED}, to distinguish AI-generated text from human-authored text. \textit{M-TRUST} models contextual variations induced by meaning-preserving linguistic transformations to identify discriminative patterns between human-authored and AI-generated texts, whereas RED models residual representation-level drift induced by these transformations to distinguish AI-generated from human-authored text. We conduct extensive experiments and ablation studies on datasets spanning linguistically diverse languages, including low-resource languages, to evaluate the effectiveness of the proposed approaches. The results reveal that our method outperforms state-of-the-art models in the literature. Furthermore, our proposed methods efficiently detect AI-generated text content across linguistically diverse languages, including low-resource languages. This study identifies multimodal AI-generated content detection, encompassing text, images, audio, and video, as future work.


\section*{Declaration of Generative AI Use}
Generative AI tools were used to assist with language refinement, grammatical corrections, manuscript readability, and formatting of figures and tables. All AI-assisted content was reviewed and verified by the authors, who retain full responsibility for the accuracy, originality, scientific integrity, experimental results, interpretations, and conclusions presented in the manuscript.



\bibliographystyle{unsrt}  
\bibliography{references}


%
\end{document}